\documentclass[conference]{IEEEtran}
\IEEEoverridecommandlockouts
\usepackage{cite}
\usepackage{amsmath,amssymb,amsfonts}
\usepackage{algorithmic}
\usepackage[frozencache=true,cachedir=.]{minted} \usepackage{listings}

\usepackage{orcidlink}
\usepackage{fontawesome5}
\definecolor{orcidgreen}{HTML}{A6CE39}
\usepackage{hyperref}
\usepackage{graphicx}
\usepackage{siunitx}
\usepackage{textcomp}
\usepackage{xcolor}
\def\BibTeX{{\rm B\kern-.05em{\sc i\kern-.025em b}\kern-.08em
    T\kern-.1667em\lower.7ex\hbox{E}\kern-.125emX}}
\begin{document}

\title{FuzzyLogic.jl: a Flexible Library for Efficient and Productive Fuzzy Inference

}

\author{\IEEEauthorblockN{1\textsuperscript{st} Luca Ferranti}
\IEEEauthorblockA{\textit{University of Vaasa}\\
Vaasa, Finland \\
\href{https://orcid.org/0000-0001-5588-0920}{{\color{orcidgreen}\faOrcid{}}0000-0001-5588-0920}
}
\and
\IEEEauthorblockN{2\textsuperscript{nd} Jani Boutellier}
\IEEEauthorblockA{\textit{University of Vaasa}\\
Vaasa, Finland}
}

\maketitle

\begin{abstract}
This paper introduces \textsc{FuzzyLogic.jl}, a Julia library to perform fuzzy inference. The library is fully open-source and released under a permissive license. The core design principles of the library are: user-friendliness, flexibility, efficiency and interoperability. Particularly, our library is easy to use, allows to specify fuzzy systems in an expressive yet concise domain specific language, has several visualization tools, supports popular inference systems like Mamdani, Sugeno and Type-2 systems, can be easily expanded with custom user settings or algorithms and can perform fuzzy inference efficiently. It also allows reading fuzzy models from other formats such as Matlab .fis, FCL or FML. In this paper, we describe the library main features and benchmark it with a few examples, showing it achieves significant speedup compared to the Matlab fuzzy toolbox.
\end{abstract}

\begin{IEEEkeywords}
Fuzzy inference, library, toolbox, Julia, Mamdani, Sugeno, Type-2, open-source software
\end{IEEEkeywords}

\section{Introduction}
\label{sec:intro}

Since their introduction by Zadeh in 1965 \cite{zadeh1965fuzzy}, fuzzy sets and fuzzy logic have steadily gained popularity as a technique to model uncertainty. They have been successfully applied to several engineering domains such as control theory \cite{tang2001optimal, carvajal2000fuzzy}, image processing \cite{russo1996fuzzy, pal1992fuzzy} and data analysis \cite{melin2014review}. During the past few decades, different inference systems have been proposed, including Mamdani \cite{mamdani1975experiment}, Sugeno \cite{takagi1985fuzzy}, Tsukamoto \cite{tsukamoto1979approach} and more recently Type-2 \cite{karnik1999type} inference systems. With their growth in engineering and research communities, needs for a common framework to easily implement and compare different inference systems and algorithms have emerged.  

In this paper, we introduce \textsc{FuzzyLogic.jl} a library for fuzzy inference written in Julia. The core design principles of the library are

\begin{itemize}
    \item \textbf{Features:} The library supports both type-1 and type-2 Mamdani and Sugeno inference systems, and offers out-of-the-box several popular membership functions, defuzzification algorithms and T-norms.
    \item \textbf{Expressive:} The library comes with its own domain specific language (DSL) that allows to describe inference systems. 
    \item \textbf{Compatible:} The library can read fuzzy systems from popular description languages, including the IEC 61131-7 Fuzzy Control Language (FCL) \cite{IEC61131},  IEEE 1855-2016 Fuzzy Markup Language (FML) \cite{acampora2005fuzzy} and MatLab fuzzy systems format. This allows seamless integration with previously developed models without code rewriting.
    \item \textbf{Productive:} The library offers visualization and debugging tools that allows the user to tune their model in an interactive manner.
    \item \textbf{Efficient:} The library achieves state-of-the-art performance. Leveraging the Julia type system and multiple dispatch, this can be achieved without giving up on the dynamic nature of the library. Furthermore, the library allows to synthesize stand-alone optimized code.
    \item \textbf{Expandable:} The library is developed in a modular way, allowing the user to easily expand with, e.g., custom membership functions, T-norms or defuzzifiers. As we show in Section \ref{sec:examples}, it is also possible to easily craft non-standard types of systems with custom evaluation.
\end{itemize}

Summarizing, the main contribution of this paper is the introduction of FuzzyLogic.jl, a fully open-source library for fuzzy inference, released under a permissive license available at \url{https://github.com/lucaferranti/FuzzyLogic.jl}. We show how our library covers several popular features of fuzzy systems and offers an expressive DSL to synthetically yet clearly describe fuzzy inference systems. We show how our library can be extended to easily implement custom inference systems and achieve state-of-the-art performance by using three popular real-world examples as test cases.  

The paper is structured as follows: in Section II we briefly review the status of existing fuzzy inference tools and their main features. In Section III we analyze in more details the features of FuzzyLogic.jl and how it achieves its design principles. In Section IV we demonstrate how to use our library with three benchmark examples, comparing its features and performance to the Matlab Fuzzy Logic toolbox. Finally, conclusions are drawn in Section V.

\section{Related work}
In this section we briefly review existing libraries for fuzzy inference.  A systematic literature review of open-source fuzzy systems software can be found in \cite{fedez2015}. We do not intend to evaluate libraries over each other, the purpose of this section is simply to offer an unbiased review of features in other existing libraries. 

The number of software frameworks for leveraging fuzzy logic in expert systems or for fuzzy inference steeply increased in the 1980s \cite{pan1998fuzzyshell}, introducing works such as FLOPS \cite{buckley1986fuzzy}. Later frameworks, released in the 1990's and 2000's, include NEFCLASS~\cite{nauck1995nefclass}, FuzzyShell~\cite{pan1998fuzzyshell}, FuzzyCLIPS~\cite{orchard2004fuzzyclips}, FuzzyJess~\cite{orchard2001fuzzy} and XFuzzy~\cite{lopez1998xfuzzy}. XFuzzy featured a graphical user interface for designing fuzzy systems based on the XFL language, and allowed synthesizing the developed systems into C or VHDL \cite{lopez1998xfuzzy}. FuzzyCLIPS~\cite{orchard2004fuzzyclips}, Java-based FuzzyJess~\cite{orchard2001fuzzy} and FuzzyShell~\cite{pan1998fuzzyshell}, on the other hand, provided a system shell based interface mainly for describing fuzzy expert systems. Finally, NEFCLASS~\cite{nauck1995nefclass} concentrated on neuro-fuzzy learning and classification. In a sense closely related to FuzzyLogic.jl, FFLL~\cite{zarozinski2006ai} proposed a fast library for fuzzy logic written in C language.

One of the most advanced open-source fuzzy libraries is jFuzzyLogic \cite{cingolani2013jfuzzylogic}, a Java library that supports both Mamdani and Sugeno type-1 inference systems and supports FCL parsing. It also allows the user to add custom functions, such as using custom membership functions or T-norms. While the library does not directly support FML, compatibility could be achieved combining jFuzzyLogic with JFML \cite{soto2018jfml}, a Java library to parse fuzzy systems specifications compliant with the IEEE 1855-2016 standard. Based on the documentation, jFuzzyLogic does not seem to support Type-2 inference systems. The Python library PyIT2FLS \cite{haghrah2019pyit2fls} supports both Type-1 and Type-2 Mamdani and Sugeno systems, however it does not support reading FCL or FML.

If not restricted to open-source, worth mentioning is the Matlab Fuzzy Logic toolbox, which allows to design Mamdani and Sugeno systems, both Type-1 and Type-2, has some visualization tools and supports custom functions. The toolbox, however, uses its own \textsf{.fis} format to store the implemented models and does not support reading/writing FCL nor FML.

\section{Library design and features}
In this section we describe the architecture and design choices of the library, highlighting how these allowed to achieve the design goals presented in Section \ref{sec:intro}. The purpose of this section is to simply highlight the general design principles. For detailed tutorials and complete list of supported features, we refer to the library documentation \cite{ferranti2023doc}.

As higher-level ambition, the library aims to offer a general framework for fuzzy inference. In addition to offering out-of-the-box a rich repertoire of features, the library should allow to expand existing inference systems with custom algorithms or even allow users to define non-standard inference systems with minimal ad-hoc manual tuning. This allows researchers and industry professionals to easily combine and compare different algorithms. Such flexibility, however, should be achieved without sacrificing speed. Last but not least, the library should make developing inference systems as enjoyable as possible for the users. That is, the code to develop an inference system should be easy to write and easy to read.

To achieve these goals, we chose to implement our library in Julia \cite{bezanson2017julia}. This choice was mainly driven by two factors:

\begin{itemize}
    \item Julia has powerful metaprogramming features like syntactic macros (similar to Lisp). This allows us to define a custom Embedded Domain Specific Language (EDSL), to express inference systems in a Julia-like syntax.
    \item Julia type system and multiple dispatch functionalities were initially designed to achieve both flexibility and speed. In our library, each algorithm has its own type and the full inference pipeline is written in a generic and modular way. This makes the inference system flexible and allows the users to easily replace each component with their custom implementation. Furthermore, the type of the full inference system has its individual components (T-norm algorithm, defuzzification algorithm, etc.) as type parameters. Leveraging the Julia type inference system, the compiler is already able to determine what algorithm to use at compile time, practically removing any computational overhead of the generic implementation from runtime.
\end{itemize}

\subsection{Domain-Specific-Language}

The library offers an embedded domain specific language to describe fuzzy inference systems in an expressive yet concise manner. As an example of the language, Listing \ref{code:tipper} shows how to implement the tipper inference system.

The EDSL is implemented exploiting Julia metaprogramming capabilities such as syntactic macros. In the example above, the macro \textsf{@mamfis} takes as input the Julia code, in the form of a parsed syntax tree and produces new Julia code, which is then evaluated. Inside the macro, the syntax tree can be traversed and transformed to produce new code. This allows to create embedded domain specific languages that preserve the Julia syntax, but have completely different semantics (meaning). For example, the first line of the code, while looking syntactically as a function definition,  should actually be read as ``create a fuzzy system called tipper, which has input variables called food and service and an output variable called tip". This allows us to program the fuzzy system with a clear Julia-like syntax, while still being able to execute the code.

\begin{listing}[t]
\begin{minted}[
%	linenos,
	fontsize=\scriptsize,
	frame = single,
        obeytabs=true,
        tabsize=2,
	]{julia}
fis = @mamfis function tipper(service, food)::tip
  service := begin
    domain = 0:10
    poor = GaussianMF(0.0, 1.5)
    good = GaussianMF(5.0, 1.5)
    excellent = GaussianMF(10.0, 1.5)
  end

  food := begin
    domain = 0:10
    rancid = TrapezoidalMF(-2, 0, 1, 3)
    delicious = TrapezoidalMF(7, 9, 10, 12)
  end

  tip := begin
    domain = 0:30
    cheap = TriangularMF(0, 5, 10)
    average = TriangularMF(10, 15, 20)
    generous = TriangularMF(20, 25, 30)
  end

  service == poor || food == rancid --> tip == cheap
  service == good --> tip == average
  service==excellent || food==delicious --> tip==generous
end
\end{minted}
\caption{Tipper Mamdani system in FuzzyLogic.jl DSL.}\label{code:tipper}

\end{listing}

\subsection{Library API}

\subsubsection{Type system}
 To best understand the structure and type system of our library, it is instructive to look at a concrete example. We start by looking at the type of the Mamdani inference system built in the previous section, displayed in Listing \ref{code:type}.

 \begin{listing}[h!]
\begin{minted}[
	fontsize=\scriptsize,
	frame = single,
        obeytabs=true,
        tabsize=2,
	]{julia}
MamdaniFuzzySystem{MinAnd, MaxOr, MinImplication,
                   MaxAggregator, CentroidDefuzzifier,
                   FuzzyRule}
\end{minted}
\caption{Type signature of a Mamdani inference system.}\label{code:type}
\end{listing}

Each inference system has its own type, in the example above \textsf{MamdaniFuzzySystem}. The inference system has however also several type parameters (the types between braces). This means the algorithms used in each part of the inference pipeline are part of the inference system type itself. Each parameter in the type represents a setting in the inference system that can be tuned. In the case of Mamdani system, these tunable settings are

\begin{itemize}
    \item Conjuction operator
    \item Disjunction operator
    \item Implication algorithm
    \item Aggregation algorithm
    \item Defuzzification algorithm
\end{itemize}

The main advantage of having a type for each algorithm and parameterizing the whole system by its settings is in the gained efficiency. Since the algorithm is determined by the type alone, if the type is known at compile-type, the compiler can determine ahead of time, e.g., what function to use for defuzzification. This way, we keep a generic framework that supports multiple algorithms, but the choice of the algorithm is already done at compile time, removing the need to check what algorithm to use at run time.

Each setting has its own abstract type, for example \textsf{CentroidDefuzzifier} and all possible defuzzifier algorithms are subtypes of  \textsf{AbstractDefuzzifier}. This allows writing generic code and makes the API of the library easy to expand with custom functions.

As a small example to show how the framework can be expanded, let us suppose the user would like to add a custom membership function \textsf{MySingleton} with parameter $c$ defined as $\mu_c(x) = 1$ if $x = c$ and $\mu_c(x)=0$ otherwise. Such a membership function can be added to our framework with the following lines of code

 \begin{listing}[h!]
\begin{minted}[
	fontsize=\scriptsize,
	frame = single,
        obeytabs=true,
        tabsize=2,
	]{julia}
struct MySingleton{T<:Real} <: AbstractMembershipFunction
  c::T
end

(mf::MySingleton)(x) = x == mf.c ? 1 : 0 
\end{minted}
\caption{Defining a custom membership function.}\label{code:mf}

\end{listing}

In the first part, we are defining a type \textsf{MySingleton} for our membership function which has a parameter $c$. We are defining this to be a subtype of \textsf{AbstractMembershipFunction} defined in \textsf{FuzzyLogic.jl} to tell the library this should behave as a membership function. In the second part, we are defining how this membership function should be evaluated. This line of code will make objects of type \textsf{MySingleton} callable as functions.

These few lines of code are everything we need to define a new membership function, which can now be used seamlessly with all the built-in functionalities in the library. In a similar manner, we can also define custom T-norms, aggregation methods and defuzzification methods. In the next section, we will also show how we can reuse the library DSL and existing type system to design a non-standard inference system.

\subsubsection{Variables}

Variables in our library are stored in a hash-table, where each key is a symbol corresponding to the variable name and the value is an object of type \textsf{Variable}. This has two parameters, a domain and a hash-table storing the membership functions corresponding to the variable.

\subsubsection{Rules}

The atomic component of a fuzzy rule is a fuzzy relation, for example ``service is good". A fuzzy rule is then composed by an antecedent and a consequent. The antecedent is a fuzzy proposition, while the consequent is a vector of fuzzy relations. Fuzzy propositions in our library are stored lazily as a tree, where each leaf is a fuzzy relation. For example, the antecedent of the first rule of tipper is constructed internally as follows.

 \begin{listing}[h!]
\begin{minted}[
	fontsize=\scriptsize,
	frame = single,
        obeytabs=true,
        tabsize=2,
	]{julia}
FuzzyOr(
    FuzzyRelation(:service, :poor),
    FuzzyRelation(:food, :rancid)
)
\end{minted}
\caption{Internal tree representation of a fuzzy rule.}\label{code:rule}

\end{listing}

As it can be noticed, a fuzzy relation is simply a pair of symbols, representing  a variable and a membership function. Fuzzy relations can be evaluated as a function, giving as input the values of the variables and the lookup tables stored in the fuzzy system to determine what membership functions to use. Similarly, fuzzy propositions can be evaluated by recursively traversing the tree. The functions to use to evaluate the connectives can be read directly from the type of the fuzzy system, and hence are known at compile time.

\subsection{Other features}

\begin{figure}[t]
    \centering
    \includegraphics[width=\linewidth]{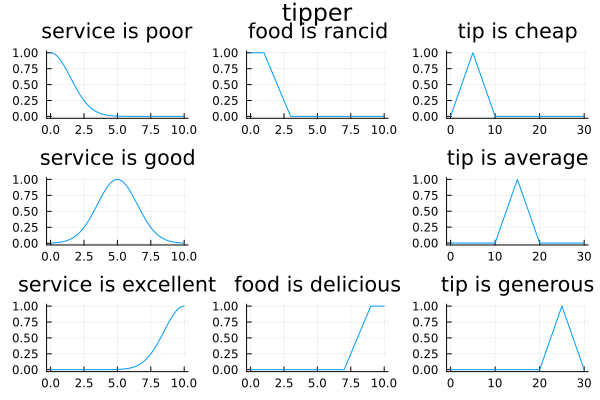}
    \caption{Graphical visualization of the tipper inference system. Automatically created by the library.}
    \label{fig:tipper}
\end{figure}
\textbf{Plotting}: In addition to the features described above. The library also offers plotting functionalities to help visualization and debugging. For example, Figure \ref{fig:tipper} shows the graphical representation of the tipper inference system created in Listing \ref{code:tipper}. This plot is automatically generated by simply calling \textsf{plot(fis)}. The plotting functionalities are implemented as generic code on the abstract types, meaning that the visualization would work out of the box, e.g., also for custom membership functions implemented by the user.

\textbf{Interoperability:} A core design principle of the library is to facilitate interoperability and minimize manual rewriting of legacy code. For this reason, the library comes with a built-in parser for IEC 61131-7 Fuzzy Control Language, supporting all the required features and several of the optional features. Furthermore, the library is also able to read fuzzy models generated with Matlab Fuzzy Logic Toolbox or models in the Fuzzy Markup Language.

\textbf{Code generation:} While the type system is designed to achieve both general flexibility and efficiency, once a fuzzy model has been designed and tuned, it may be desirable to generate stand-alone code. Currently, the library is able to compile fuzzy models to an optimized stand-alone Julia function that can be used fully independently from the library.

\section{Examples and benchmarks}
\label{sec:examples}
In this section we benchmark our library on three testcases, comparing execution time to the Matlab library. The codes to reproduce the experiments can be found at \url{https://github.com/lucaferranti/fuzzieee2023}. The first example is the tipper system presented in the previous section.

The second example is a controller for a wall-following robot presented in \cite{mucientes2009learning}. This is a Mamdani inference system with 4 input variables, each with $4$, $2$, $4$ and $2$ membership functions respectively; 2 output variables with $9$ membership functions each, and $41$ rules.
 
The third example is the fuzzy denoising algorithm presented in \cite{russo1996fuzzy}. This fuzzy inference system has $8$ input variables, all with the same membership functions, one output variable and 26 rules. This system is however slightly unconventional, namely it follows the traditional inference system pipeline only until the rule antecedent evaluation. Let

\begin{equation}\label{eq:denoise}
    \begin{split}
        &\lambda_1 = \max\{r_i: i = 1\ldots13\}\\
        &\lambda_2 = \max\{r_i: i = 14\ldots26\}\\
        &\lambda_0 = \max(0, 1 - \lambda_1 - \lambda_2)\\
        &y = (L - 1)\frac{\lambda_1 - \lambda_2}{\lambda_1 + \lambda_2 + \lambda_0},
    \end{split}
\end{equation}

where $r_i$ is the output of the antecedent of the $i$th rule, $y$ is the final output and $L$ is the number of gray levels in the original image. This fuzzy system is fairly unconventional in the sense that it uses the result of each antecedent as crisp output of the rule, partitions the rules into two groups that are aggregated separately, and computes the final result from the two aggregated outputs. The implementation of this non-standard fuzzy system is shown in Listing \ref{code:denoise}, where the first part defines the inference system. While this is not a Mamdani system, we can reuse the \textsf{@mamfis} macro, since the Mamdani system holds all the pieces we need. This example also highlights a core feature of FuzzyLogic.jl, as it can specify variables as a vector $x[1:8]$ and it supports syntactic for-loops, which can be used to avoid manual typing when we have several variables constructed in a regular manner. We could also have used a similar for-loop syntax for the rules, by computing each variable index as a function of the rule index, but this would probably have lead to less readable rules.

The function detector defined below is the custom evaluation pipeline described in Equation \ref{eq:denoise}. As mentioned in the previous section, each logical proposition is stored as a tree and can be called as a function. In the first two lines we evaluate the antecedent of each rule at the given inputs. The \textsf{fis} is passed as input to the antecedent too, as it contains in its type signature the information on how to evaluate each logical connective. After having evaluated the antecedents, we proceed with the custom inference pipeline. As this example shows, leveraging the library's flexible DSL and modular design, we could implement a non-standard inference system in only a few lines of code, hijacking the traditional inference pipeline to evaluate only part of it and build on top of it our custom inference procedure.

\begin{listing}[t]
\begin{minted}[
%	linenos,
	fontsize=\scriptsize,
	frame = single,
        obeytabs=true,
        tabsize=2,
	]{julia}
fis = @mamfis function denoise(x[1:8])::y
for i in 1:8
    x[i] := begin
        domain = -1000:1000
        POS = TriangularMF(-255.0, 255.0, 765.0)
        NEG = TriangularMF(-765.0, -255.0, 255.0)  
    end
end

y := begin
    domain = -1000:1000
    PO = TriangularMF(-255.0, 255.0, 765.0)
    NE = TriangularMF(-765.0, -255.0, 255.0)
end

x2==POS && x5 == POS && x7 == POS --> y == PO
x5==POS && x7 == POS && x4 == POS --> y == PO
x7==POS && x4 == POS && x2 == POS --> y == PO
x4==POS && x2 == POS && x5 == POS --> y == PO
x1==POS && x3 == POS && x8 == POS && x6 == POS --> y == PO
x1==POS && x2 == POS && x3 == POS && x5 == POS --> y == PO
x2==POS && x3 == POS && x5 == POS && x8 == POS --> y == PO
x3==POS && x5 == POS && x8 == POS && x7 == POS --> y == PO
x5==POS && x8 == POS && x7 == POS && x6 == POS --> y == PO
x8==POS && x7 == POS && x6 == POS && x4 == POS --> y == PO
x7==POS && x6 == POS && x4 == POS && x1 == POS --> y == PO
x6==POS && x4 == POS && x1 == POS && x2 == POS --> y == PO
x4==POS && x1 == POS && x2 == POS && x3 == POS --> y == PO
x2==NEG && x5 == NEG && x7 == NEG --> y == NE
x5==NEG && x7 == NEG && x4 == NEG --> y == NE
x7==NEG && x4 == NEG && x2 == NEG --> y == NE
x4==NEG && x2 == NEG && x5 == NEG --> y == NE
x1==NEG && x3 == NEG && x8 == NEG && x6 == NEG --> y == NE
x1==NEG && x2 == NEG && x3 == NEG && x5 == NEG --> y == NE
x2==NEG && x3 == NEG && x5 == NEG && x8 == NEG --> y == NE
x3==NEG && x5 == NEG && x8 == NEG && x7 == NEG --> y == NE
x5==NEG && x8 == NEG && x7 == NEG && x6 == NEG --> y == NE
x8==NEG && x7 == NEG && x6 == NEG && x4 == NEG --> y == NE
x7==NEG && x6 == NEG && x4 == NEG && x1 == NEG --> y == NE
x6==NEG && x4 == NEG && x1 == NEG && x2 == NEG --> y == NE
x4==NEG && x1 == NEG && x2 == NEG && x3 == NEG --> y == NE
end

function detector(fis, inputs)
  L1 = maximum(fis.rules[i].antecedent(fis, inputs)
               for i in 1:13)
  L2 = maximum(fis.rules[i].antecedent(fis, inputs)
               for i in 14:26)
  L0 = max(0, 1 - L1 - L2)
  return 255 * (L1 - L2) / (L1 + L2 + L0)
end

\end{minted}
\caption{Inference system for image denoising.}\label{code:denoise}
\end{listing}

To conclude the experiments, we implemented the presented examples in our library and report the timings in Table \ref{tab:benchmarks}. For the first and second example, the implementation was presented in Listings \ref{code:tipper} and \ref{code:denoise}. The second example was directly read by an available description in FCL format. The experiments were run on a laptop with 12th Gen Intel(R) Core(TM) i7-1255U processor and 64GB of RAM. As can be seen, our library is significantly faster than Matlab. One reason for this is that Matlab stores the parameters of the algorithm as strings and determines what algorithm to use at runtime, while our library can determine ahead of time, e.g., what defuzzifier implementation to pick, removing any need for comparisons at runtime.  
\begin{table}[t]
    \centering
    \caption{Benchmarks on selected examples}
    \label{tab:benchmarks}
    \begin{tabular}{c||cc}
         Test case& FuzzyLogic.jl &Matlab fuzzy toolbox\\\hline\hline
         Tipper&\SI{6}{\micro\second}&\SI{60}{\micro\second}\\
         Robot controller&\SI{92}{\micro\second}&\SI{300}{\micro\second}\\
         Image denoising&\SI{9}{\micro\second}&\SI{120}{\micro\second}\\
    \end{tabular}
\end{table}

\section{Conclusions}
In this paper we presented \textsc{FuzzyLogic.jl}, a Julia library for fuzzy inference. We described the main functionalities of the library and how it allows to design, visualize and debug different inference systems (Mamdani and Sugeno, both type~1 and type~2), and use them to perform efficient inference in different application domains. The library being open source and MIT licensed, it is well suited for adoption both in industry and academia. Furthermore, the library is extensible and allows users to add new algorithms or inference systems on top of the basic domain specific language and infrastructure.

\bibliographystyle{IEEEtran}
\bibliography{refs}

\end{document}